\documentclass[10pt,twocolumn,letterpaper]{article}

\usepackage[pagenumbers]{cvpr} 

\definecolor{cvprblue}{rgb}{0.21,0.49,0.74}
\usepackage[pagebackref,breaklinks,colorlinks,allcolors=cvprblue]{hyperref}
\usepackage{multirow}
\usepackage{tikz}
\usepackage{float}
\usepackage{graphicx}
\usepackage{arydshln}
\usepackage{algorithm}
\usepackage{svg}
\usepackage{algorithmic}
\usepackage{subcaption}
\usepackage{amsmath}
\DeclareMathOperator*{\argmax}{arg\,max}
\usepackage{colortbl}

\definecolor{aliceblue}{rgb}{0.94, 0.97, 1.0}
\definecolor{lightgray}{rgb}{0.95, 0.95, 0.95}

\newcommand{\model}{SAMURAI}
\title{SAMURAI: Adapting Segment Anything Model \\for Zero-Shot Visual Tracking with Motion-Aware Memory}

\author{Cheng-Yen Yang \quad
Hsiang-Wei Huang \quad
Wenhao Chai \quad
Zhongyu Jiang \quad 
Jenq-Neng Hwang \vspace{4pt}\\  University of Washington \vspace{4pt} \\
{\tt\small \{cycyang, hwhuang, wchai, zyjiang, hwang\} @ uw.edu}}

\begin{document}
\maketitle
\begin{abstract}
The Segment Anything Model 2 (SAM~2) has demonstrated strong performance in object segmentation tasks but faces challenges in visual object tracking, particularly when managing crowded scenes with fast-moving or self-occluding objects. Furthermore, the fixed-window memory approach in the original model does not consider the quality of memories selected to condition the image features for the next frame, leading to error propagation in videos. This paper introduces SAMURAI, an enhanced adaptation of SAM~2 specifically designed for visual object tracking. By incorporating temporal motion cues with the proposed motion-aware memory selection mechanism, SAMURAI effectively predicts object motion and refines mask selection, achieving robust, accurate tracking without the need for retraining or fine-tuning. SAMURAI operates in real-time and demonstrates strong zero-shot performance across diverse benchmark datasets, showcasing its ability to generalize without fine-tuning. In evaluations, SAMURAI achieves significant improvements in success rate and precision over existing trackers, with a 7.1\% AUC gain on LaSOT$_{\text{ext}}$ and a 3.5\% AO gain on GOT-10k. Moreover, it achieves competitive results compared to fully supervised methods on LaSOT, underscoring its robustness in complex tracking scenarios and its potential for real-world applications in dynamic environments. Code and results are available at \href{https://github.com/yangchris11/samurai}{https://github.com/yangchris11/samurai}.
\end{abstract}    
\section{Introduction}
\label{sec:intro}

Segment Anything Model~(SAM)~\cite{kirillov2023segment} has demonstrated impressive performance in segmentation tasks. Recently, SAM 2~\cite{ravi2024sam} incorporates a streaming memory architecture, which enables it to process video frames sequentially while maintaining context over long sequences. While SAM~2 has shown remarkable capabilities in Video Object Segmentation (VOS~\cite{xu2018youtube}) tasks, generating precise pixel-level masks for objects throughout a video sequence, it still faces challenges in Visual Object Tracking (VOT~\cite{roffo2016visual}) scenarios. 

\begin{figure*}[t]
    \centering
    \vspace{-60pt}
    \includegraphics[width=0.98\linewidth]{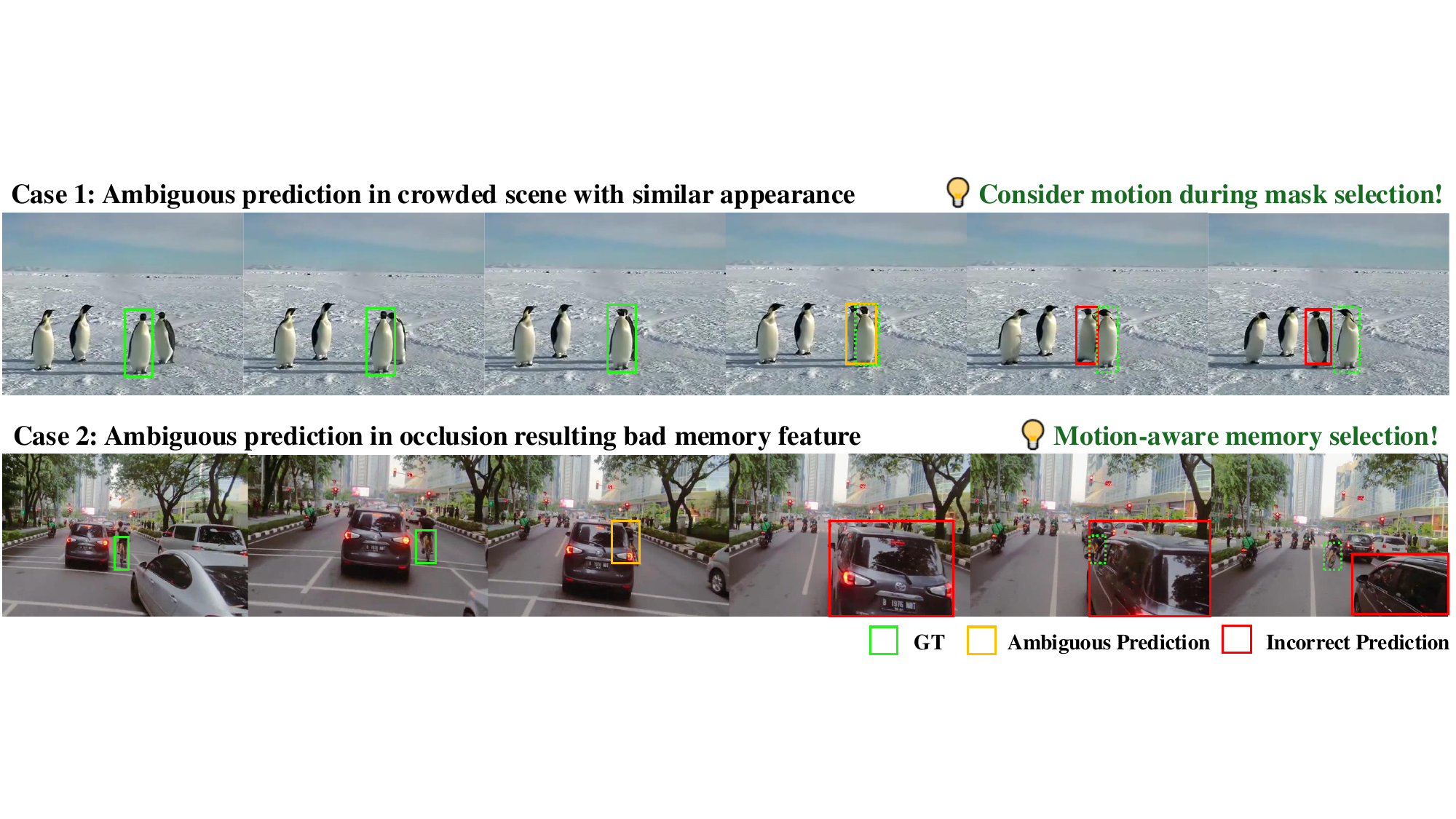}
    \vspace{-60pt}
    \caption{Illustration of two common failure cases in visual object tracking using SAM 2: (1) In a crowded scene with similar appearances between target and background objects, SAM 2 tends to ignore the motion cue and predict where the mask has the higher IoU score. (2) The original memory bank simply chooses and stores the previous $n$ frames into the memory bank, resulting in introducing some bad features during occlusion.} 
    \label{fig:intro}
\end{figure*}

The primary concern in VOT is maintaining consistent object identity and location despite occlusions, appearance changes, and the presence of similar objects. However, SAM 2 often neglects motion cues when predicting masks for subsequent frames, leading to inaccuracies in scenarios with rapid object movement or complex interactions. This limitation is particularly evident in crowded scenes, where SAM 2 tends to prioritize appearance similarity over spatial and temporal consistency, resulting in tracking errors. As illustrated in Figure~\ref{fig:intro}, there are two common failure patterns: confusion in crowded scenes and ineffective memory utilization during occlusions. 

To address these limitations, we propose incorporating motion information into SAM 2's prediction process. By leveraging the history of object trajectories, we can enhance the model's ability to differentiate between visually similar objects and maintain tracking accuracy in the presence of occlusions. Additionally, optimizing SAM 2's memory management is crucial. The current approach~\cite{ravi2024sam,ding2024sam2long} of indiscriminately storing recent frames in the memory bank introduces irrelevant features during occlusions, compromising tracking performance. Addressing these challenges is essential to adapt SAM 2's rich mask information for robust video object tracking.

To this end, we propose \textbf{SAMURAI}, a \textbf{\underline{SAM}}-based \textbf{\underline{U}}nified and \textbf{\underline{R}}obust zero-shot visual tracker with motion-\textbf{\underline{A}}ware \textbf{\underline{I}}nstance-level memory. Our proposed method incorporates two key advancements: (1) a motion modeling system that refines the mask selection, enabling more accurate object position prediction in complex scenarios, and (2) an optimized memory selection mechanism that leverages a hybrid scoring system, combining the original mask affinity, object, and motion scores to retain more relevant historical information, so as to enhance the model's overall tracking reliability.

In conclusion, this paper makes the following contributions:

\begin{itemize}[leftmargin=7.5mm]
\setlength{\itemsep}{2pt}
    \item We enhance the visual tracking accuruacy of SAM~2 by incorporating motion information through \textbf{motion modeling}, to effectively handle the fast-moving and occluded objects.
    \item We proposed a \textbf{motion-aware memory selection} mechanism that reduces error in crowded scenes in contrast to the original fixed-window memory by selectively storing relevant frames decided by a mixture of motion and affinity scores.
    \item Our zero-shot SAMURAI achieves state-of-the-art performance on LaSOT, LaSOT$_{\text{ext}}$, GOT-10k, and other VOT benchmarks without additional training or fine-tuning, demonstrating strong generalization of our proposed modules across diverse datasets.
\end{itemize}

\section{Related Works}
\label{sec:related_works}

\subsection{Visual Object Tracking (VOT)}
Visual Object Tracking~(VOT)~\cite{roffo2016visual} aims to track objects in challenging video sequences that include variations in object scale, occlusions, and complex backgrounds so as to elevate the robustness and accuracy of tracking algorithms. Siamese-based~\cite{chen2020siamese,zhang2019deeper} and transformer-based~\cite{cui2022mixformer,yan2021learning} trackers are common approaches by learning embedding similarity. However, due to lacking self-correction of these trackers in the single forward pass evaluation scheme, they can easily drift toward distractors. To this end, recent works~\cite{yang2018learning,fu2021stmtrack} further introduce memory bank and attention to find a better mapping between current frame and history information.

\subsection{Segment Anything Model (SAM)}

The Segment Anything Model (SAM)~\cite{kirillov2023segment} has sparked considerable follow-up research since its introduction. SAM introduces a prompt-based segmentation approach, where users could input points, bounding boxes, or text to guide the model in segmenting any object within an image. The use of SAM has wide-ranging applications like in video understanding~\cite{chai2024auroracap,song2024moviechat,song2024moviechat+} and editing~\cite{chai2023stablevideo}. Since then, various works have built upon SAM. For example, SAM~2~\cite{ravi2024sam} expands the model's capabilities to video segmentation~\cite{cheng2023tracking}, incorporating memory mechanisms for tracking objects across multiple frames in dynamic video sequences. Additionally, efforts have been made to create more efficient variants of SAM for resource-constrained environments, aiming to reduce its computational demands~\cite{xiong2024efficientsam, zhao2023fast}. Research in medical imaging has also adopted SAM for specialized tasks~\cite{ma2024segment}. Recently, SAM2Long~\cite{ding2024sam2long} uses tree-based memory to enhance object segmentation for long video. However, their higher FPS video sequences and deeper memory tree architectures require exponentially more computing power and memory storage due to the overhead of storing exact paths and time-constrained memory paths. On the other hand, our proposed SAMURAI model, which is built upon SAM~2,  has been trained on large-scale segmentation datasets to overcome these challenges and ensure good performance and generalization ability. 

\subsection{Motion Modeling}
Motion modeling is an important component in tracking tasks, which can be categorized into heuristic and learnable approaches. Heuristic methods, such as the widely-used Kalman Filter (KF)~\cite{kalman1960new}, rely on fixed motion priors and predefined hyper-parameters to predict object trajectories. While KF has proven effective in many tracking benchmarks, it often fails in scenarios with intense or abrupt motion. Other methods~\cite{aharon2022bot} attempt to counteract intense or abrupt object motion by compensating for camera movement before applying traditional KF-based prediction. However, both the standard and noise scale adaptive (NSA) Kalman Filters~\cite{du2021giaotracker} come with a multitude of hyper-parameters, potentially restricting their effectiveness to specific types of motion scenarios. In contrast, learnable motion models have attracted increasing interest due to their data-driven nature. Tracktor~\cite{bergmann2019tracking} is the first to use trajectory boxes as Regions of Interest (RoI) in a Faster-RCNN to extract features and regress the object’s position across frames.
MotionTrack~\cite{xiao2024motiontrack} enhances tracking by learning past trajectory representations to predict future movements. MambaTrack~\cite{huang2024exploring} further explores different learning-based motion models architecture like transformer~\cite{vaswani2017attention} and state-space model~(SSM)~\cite{gu2023mamba}. Our approach is also a learning-based motion modeling with an enhanced heuristic scheme.

\begin{figure*}[t]
    \centering
    \includegraphics[width=0.95\linewidth]{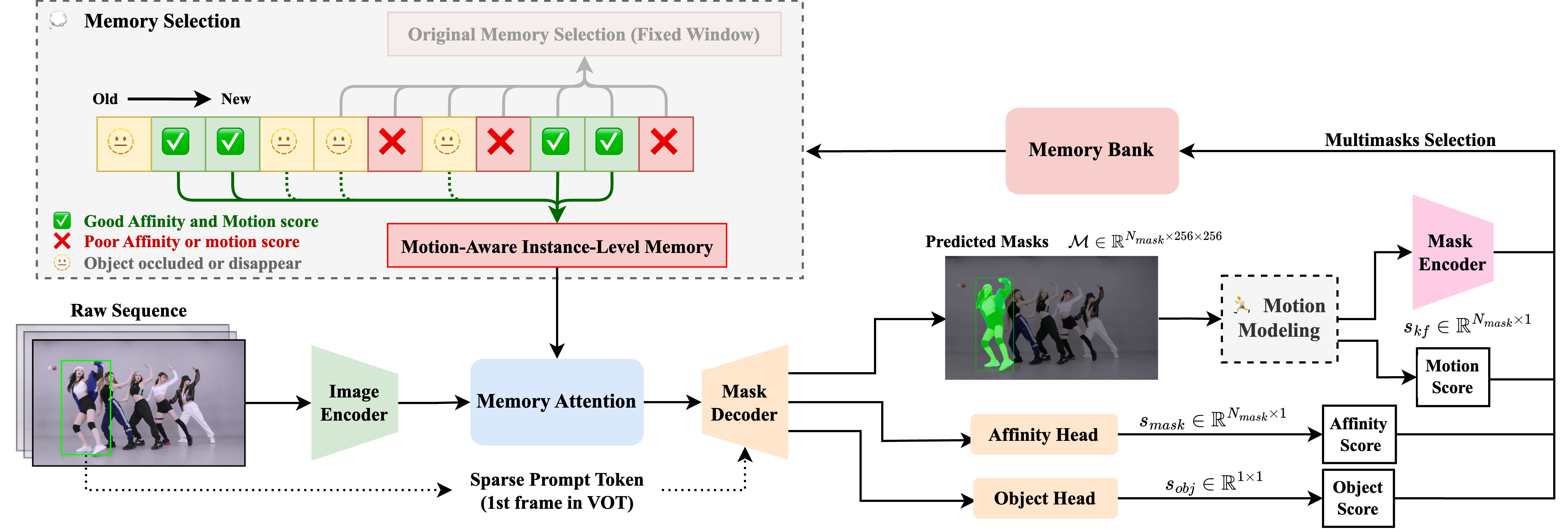}
    \caption{The overview of our SAMURAI visual object tracker.} 
    \label{fig:pipeline}
\end{figure*}

\section{Revisiting Segment Anything Model 2}

Segment Anything Model 2 \cite{ravi2024sam2} contains (1) an image encoder, (2) a mask decoder with a prompt encoder, (3) a memory attention layer, and (4) a memory encoder. We will introduce some preliminaries of SAM 2 and specifically point out the part where SAMURAI is being added.

\paragraph{Prompt Encoder.} The prompt encoder design follows SAM \cite{ravi2024sam}, in which it takes two types of prompts, including sparse (e.g., points, bounding boxes) and dense (e.g., masks). The prompt tokens output by the prompt encoder can be represented as $x_{prompt} \in N_{tokens} \times d$. In the visual object tracking, where the ground-truth bounding box of the target object of the first frame $t_0$ is provided, SAM 2 takes the positional encoding for the top-left and bottom-right points as inputs while the rest of the sequence uses the predicted mask $\bar{\mathcal{M}}_{t-1}$ from the previous frame as the input to the prompt encoder.

\paragraph{Mask Decoder.} The memory decoder is designed to take the memory-conditioned image embeddings produced by the memory attention layer along with the prompt tokens from the prompt encoder as its inputs.  Its multi-head branches can then generate a set of predicted masks, along with the corresponding mask affinity score $s_{mask}$ (it is referred to as IoU score in \cite{ravi2024sam, ravi2024sam2}), and one object score $s_{obj}$ for the frame as outputs.

\begin{equation}
    \mathbb{M} = \{(\mathcal{M}_0, s_{mask,0}), (\mathcal{M}_1, s_{mask,1}), \dots\}.
\end{equation}

The affinity mask score prediction of SAM 2 is supervised with MAE loss as it can represent the overall confidence of the mask, while the object prediction is supervised with cross-entropy loss to determine whether a mask should exist in the frame or not. In the original implementation, the final output mask, $\bar{\mathcal{M}} = \mathcal{M}_i$, is selected based on the highest affinity score among the $N_{mask}$ output masks. 

\begin{equation}
    i = \argmax_{i \in [0, N_{mask}-1]} s_{mask,i}  \quad \text{where} \quad s_{obj,i} > 0 
\end{equation}

However, the affinity score is not a very robust indicator in the case of visual tracking, especially in crowded scenarios where similar objects self-occlude with each other. We introduce an extra \textbf{motion modeling} to keep track of the motion of the target and provide an additional motion score to aid the selection of the prediction.

\paragraph{Memory Attention Layer.} The Memory attention block first performs self-attention with the frame embeddings and then performs cross-attention between the image embeddings and the contents of the memory bank. The unconditional image embeddings, therefore, get contextualized with the previous output masks, previous input prompts, and object pointers.

\paragraph{Memory Encoder and Memory Bank.} After the mask decoder generates output masks, the output mask is passed through a memory encoder to obtain a memory embedding. A new memory is created after each frame is processed. These memory embeddings are appended to a Memory Bank, which is a first-in-first-out (FIFO) queue of the latest memories generated during video decoding. At any given time $t$ in the sequence, we can form the memory bank $B_t$ as:

\begin{equation}
    B_t = [ m_{t-1}, m_{t-2}, \dots, m_{t-N_{mem}} ]
\end{equation}

\noindent which takes the past $N_{mem}$ frames' output $m$ as the components of the memory bank. 

This straightforward fixed-window memory implementation may suffer from encoding the incorrect or low-confidence object, which will cause the error to propagate considerably when in the context of a long sequence visual tracking task. Our proposed \textbf{motion-aware memory selection} will replace the original memory bank composition to ensure that better memory features can be kept and conditioned onto the image feature.

\section{Method}
\label{sec:method}

SAM~2 has demonstrated strong performance in basic Visual Object Tracking (VOT) and Video Object Segmentation (VOS) tasks. However, the original model can mistakenly encode incorrect or low-confidence objects, leading to substantial error propagation in long-sequence VOT.

To address the above issues, we propose a Kalman Filter (KF)-based motion modeling on top of the multi-masks selection (in \ref{subsec:motion}) and an enhanced memory selection based on a hybrid scoring system that combines affinity and motion scores (in \ref{subsec:memory}). These enhancements are designed to strengthen the model’s ability to track objects accurately in complex video scenarios. Importantly, this approach does not require fine-tuning, nor does it require additional training,  and it can be integrated directly into the existing SAM 2 model. By improving the selection of predicted masks without additional computational overhead, this method provides a reliable, real-time solution for online VOT.

\subsection{Motion Modeling}
\label{subsec:motion}

Motion modeling has long been an effective approach to Visual Object Tracking~(VOT) and Multiple Object Tracking~(MOT)~\cite{zhang2022bytetrack, cao2023observation, aharon2022bot} in resolving association ambiguities. We employ the linear-based Kalman filter~\cite{kalman1960new} as our baseline to demonstrate the incorporation of motion modeling in improving tracking accuracy. 

In our visual object tracking framework, we integrate the Kalman filter to enhance bounding box position and dimension predictions, which in turn helps select the most confident mask out of $N$ candidates from $\mathcal{M}$. We define the state vector $\boldsymbol{x}$ as:

\begin{equation}
\boldsymbol{x} = [x, y, w, h, \dot{x}, \dot{y}, \dot{w}, \dot{h}]^T
\end{equation}

\noindent where {\(x\), \(y\)} represents the center coordinate of the bounding box, \(w\) and \(h\) denote its width and height, respectively, and their corresponding velocities are represented by the dot notation. For each mask $\mathcal{M}_i$, the corresponding bounding box $\boldsymbol{d}_i$ is derived by computing the minimum and maximum $x$ and $y$ coordinates of the mask's non-zero pixels. The Kalman filter operates in a predict-correct cycle, where the state prediction $\boldsymbol{\hat{x}}_{t+1|t}$ is given by:

\begin{equation}
\boldsymbol{\hat{x}}_{t+1|t} = \boldsymbol{F\hat{x}}_{t|t},
\end{equation}

\begin{equation}
    s_{kf} = IoU(\boldsymbol{\hat{x}}_{t+1|t}, \mathcal{M})
\end{equation}

\noindent the KF-IoU score $s_{kf}$ is then computed by calculating the Intersection over Union (IoU) between the predicted masks $\mathcal{M}$ and the bounding box derived from the Kalman filter's predicted state. We then select the mask that maximizes a weighted sum of the KF-IoU score and the original affinity score:
\begin{equation}
\mathcal{M}^* = \argmax_{\mathcal{M}_i} (\alpha_{kf} \cdot s_{kf}(\mathcal{M}_i) + (1-\alpha_{kf}) \cdot s_{mask}(\mathcal{M}_i)).
\end{equation}

Finally, the update is performed using:

\begin{equation}
\boldsymbol{\hat{x}}_{t|t} = \boldsymbol{\hat{x}}_{t|t-1} + \boldsymbol{K}_t(\boldsymbol{z}_t - \boldsymbol{H\hat{x}}_{t|t-1})
\end{equation}

\noindent where $\boldsymbol{z}_t$ is the measurement, the bounding box derived from the mask we selected, used to update. \(\boldsymbol{F}\) is the linear state transition matrix, \(\boldsymbol{K}_n\) is the Kalman gain, and \(\boldsymbol{H}\) is the observation matrix. Furthermore, to ensure the robustness of the motion modeling after the targeted object reappears or the poor mask qualities for a certain period of time, we also maintain a stable motion state where we take consideration of the motion module if and only if the tracked object is being successfully update in the past $\tau_{kf}$ frames. 

\subsection{Motion-Aware Memory Selection}
\label{subsec:memory}

\begin{table*}[tbh]
    \small
    \centering
        \caption{Visual object tracking results on \textbf{LaSOT} \cite{fan2019lasot}, \textbf{LaSOT$_{\text{ext}}$} \cite{fan2021lasotext}, and \textbf{GOT-10k} \cite{huang2019got}. \textsuperscript{$\dagger$} LaSOT$_\text{ext}$ are evaluated on trackers to be trained with LaSOT.  \textsuperscript{$\ddagger$} GOT-10k protocol only allows trackers to be trained using its corresponding train split. The T, S, B, L represents the size of the ViT-based backbone while the subscript is the search region. \textbf{Bold} represents the best while \underline{underline} represents the second.}
    \resizebox{0.97\linewidth}{!}{
    \begin{tabular}{l|c|ccc|ccc|ccc}
        \toprule
        \multirow{2}{*}{Trackers} & \multirow{2}{*}{Source} & \multicolumn{3}{c|}{LaSOT} & \multicolumn{3}{c|}{LaSOT$_{\text{ext}}$\textsuperscript{$\dagger$}} & \multicolumn{3}{c}{GOT-10k \textsuperscript{$\ddagger$}}  \\ 
         & & AUC(\%) & P$_{\text{norm}}$(\%) & P(\%) & AUC(\%) & P$_{\text{norm}}$(\%) & P(\%) & AO(\%) & OP$_{0.5}$(\%) & OP$_{0.75}$(\%) \\ \midrule
         \multicolumn{1}{l|}{\textit{Supervised method}} &  &  &  &  &  &  & & &\\
         SiamRPN++ \cite{li2019siamrpn++} & CVPR'19 & 49.6 & 56.9 & 49.1 & 34.0 & 41.6 & 39.6 & 51.7 & 61.6 & 32.5 \\
         DiMP$_{288}$ \cite{danelljan2020dimp} & CVPR'20 & 56.3 & 64.1 & 56.0 & - & - & - & 61.1 & 71.7 & 49.2\\
         TransT$_{256}$ \cite{chen2021transt} & CVPR'21 & 64.9 & 73.8 & 69.0 & - & - & - & 67.1 & 76.8 & 60.9\\
         AutoMatch$_{255}$ \cite{zhang2021automatch} & ICCV'21 &  58.2 & 67.5 & 59.9 & - & - & - & 65.2 & 76.6 & 54.3 \\
         STARK$_{320}$ \cite{yan2021stark} & ICCV'21 & 67.1 & 76.9 & 72.2 & - & - & - & 68.8 & 78.1 & 64.1 \\
         SwinTrack-B$_{384}$ \cite{lin2022swintrack} & NeurIPS'22 & 71.4 & 79.4 & 76.5 & - & - & -  & 72.4 & 80.5 & 67.8\\
         MixFormer$_{288}$ \cite{cui2022mixformer} & CVPR'22 & 69.2 & 78.7 & 74.7 & - & - & - & 70.7 & 80.0 & 67.8\\
         OSTrack$_{384}$ \cite{ye2022ostrack} & ECCV'22 & 71.1 & 81.1 & 77.6 & 50.5 & 61.3 & 57.6 & 73.7 & 83.2 & 70.8 \\ 
         ARTrack-B$_{256}$ \cite{wei2023artrack} & CVPR'23 & 70.8 & 79.5 & 76.2 & 48.4 & 57.7 & 53.7 & 73.5 & 82.2 & 70.9 \\
         SeqTrack-B$_{384}$ \cite{chen2023seqtrack} & CVPR'23 & 71.5 & 81.1 & 77.8 & 50.5 & 61.6 & 57.5 & 74.5 & 84.3 & 71.4 \\
         GRM-B$_{256}$ \cite{gao2023grm} & CVPR'23 & 69.9 & 79.3 & 75.8 & - & - &- & 73.4 & 82.9 & 70.4 \\
         ROMTrack-B$_{256}$ \cite{cai2023romtrack} & ICCV'23 & 69.3 & 78.8 & 75.6 & 47.2 & 53.5 & 52.9 & 72.9 & 82.9 & 70.2 \\
         TaMOs-B$_{384}$ \cite{mayer2024tamos} & WACV'24 & 70.2 & 79.3 & 77.8 & - & - & -  & - & - & -\\
         EVPTrack-B$_{384}$ \cite{shi2024evptrack} & AAAI'24 & 72.7 & 82.9 & 80.3 & 53.7 & 65.5 & 61.9 & 76.6 & 86.7 & 73.9 \\
         ODTrack-B$_{384}$ \cite{zheng2024odtrack} & AAAI'24 & 73.2 & 83.2 & 80.6 & 52.4 & 63.9 & 60.1 & 77.0 & 87.9 & 75.1 \\
         ODTrack-L$_{384}$ \cite{zheng2024odtrack} & AAAI'24 & \underline{74.0} & \textbf{84.2} & \textbf{82.3} & 53.9 & 65.4 & 61.7 & 78.2 & 87.2 & \textbf{77.3} \\
         HIPTrack-B$_{384}$ \cite{cai2024hiptrack} & CVPR'24 & 72.7 & 82.9 & 79.5 & 53.0 & 64.3 & 60.6 & 77.4 & 88.0 & 74.5\\
         AQATrack-B$_{256}$ \cite{xie2024aqatrack} & CVPR'24 & 71.4 & 81.9 & 78.6 & 51.2 & 62.2 & 58.9 & 73.8 & 83.2 & 72.1 \\
         AQATrack-L$_{384}$ \cite{xie2024aqatrack} & CVPR'24 & 72.7 & 82.9 & 80.2 & 52.7 & 64.2 & 60.8 & 76.0 & 85.2 & 74.9\\
         LoRAT-B$_{224}$ \cite{lin2025lorat} & ECCV'24 & 71.7 & 80.9 & 77.3 & 50.3 & 61.6 & 57.1 & 72.1 & 81.8 & 70.7 \\ 
         LoRAT-L$_{224}$ \cite{lin2025lorat} & ECCV'24 & \textbf{74.2} & \underline{83.6} & \underline{80.9} & 52.8 & 64.7 & 60.0 & 75.7 & 84.9 & 75.0 \\
         \midrule
         \multicolumn{1}{l|}{\textit{Zero-shot method}} &  &  &  &  &  &  & & & \\
        \rowcolor{aliceblue} SAMURAI-T & Ours & 69.3 & 76.4 & 73.8  & 55.1 & 65.6 & 63.7 & 79.0 & 89.6 & 72.3 \\         
         \rowcolor{aliceblue}SAMURAI-S & Ours & 70.0  & 77.6 & 75.2  & \underline{58.0}  & \underline{69.6} & \underline{67.7}  & 78.8 & 88.7 & 72.9 \\ 
         \rowcolor{aliceblue}SAMURAI-B & Ours &70.7 & 78.7  & 76.2 & 57.5  & 69.3 & 67.1 & \underline{79.6}& \underline{90.8} & 72.9  \\
         \rowcolor{aliceblue} SAMURAI-L & Ours & \textbf{74.2} & 82.7 & 80.2 & \textbf{61.0} & \textbf{73.9} & \textbf{72.2} & \textbf{81.7} & \textbf{92.2} & \underline{76.9}\\
         \bottomrule
    \end{tabular}
    }
    \label{tab:lasot}
\end{table*}

The original SAM 2 prepares the conditioned visual feature of the current frame based on selecting $N_{mem}$ from the previous frames. In \cite{ravi2024sam2}, the implementation simply selects the $N_{mem}$ most recent frames based on the qualities of the target. However, this approach has the weakness of not being able to handle longer occlusion or deformation, which is common in visual object tracking tasks.

To construct an effective memory bank of object cues considering motion, we employ a selective approach for choosing frames from previous time steps based on three scoring: the mask affinity score, object occurrence score, and motion score. We select the frame as an ideal candidate for memory if and only if all three scores meet their corresponding thresholds (e.g., $\tau_{mask}$, $\tau_{obj}$, $\tau_{kf}$). We iterate back in time from the current frame and repeat the verification. We select $N_{mem}$ memories based on the above scoring function and obtain a motion-aware memory bank $B_t$:

\begin{equation}
    B_t=\{m_i|f(s_{mask}, s_{obj}, s_{kf})=1,t-N_{max}\leq i<t\}
\end{equation}


\noindent where $N_{max}$ is the maximum number of frames to look back. The motion-aware memory bank $B_t$ is subsequently passed through the memory attention layer and then directed to mask decoder $D_{mask}$ to perform mask decoding at current timestamp. Note that we follow the $N_{mem}=7$ as the SAM~2 is trained under these specific memory bank settings.

The proposed motion modeling and memory selection module can significantly enhance visual object tracking without the need for retraining and does not add any computational overhead to the existing pipeline. It is also model-agnostic and potentially applicable to other tracking frameworks beyond SAM~2. By combining motion modeling with intelligent memory selection, we can enhance tracking performance in challenging real-world applications without sacrificing efficiency.

\section{Experiments}
\label{sec:exp}
\subsection{Benchmarks}

We evaluate the zero-shot performance of our~\model~on the following VOT benchmarks: 

\vspace{-6pt}

\paragraph{LaSOT~\cite{fan2019lasot}} is a visual object tracking dataset comprising 1,400 videos across 70 diverse object categories with an average sequence length of 2,500 frames. It is divided into training and testing sets, consisting of 1,120 and 280 sequences, respectively, with 16 training and 4 testing sequences for each category.

\vspace{-6pt}

\paragraph{LaSOT$_\text{ext}$~\cite{fan2021lasotext}} is an extension to the original LaSOT dataset, introducing an additional 150 video sequences across 15 new object categories. These new sequences are specifically designed to focus on occlusions and variations in small objects, which is more challenging, and the standard protocol is to evaluate the models trained on LaSOT directly on the LaSOT$_\text{ext}$.

\vspace{-6pt}

\paragraph{GOT-10k~\cite{huang2019got}} comprises over 10,000 video segments of real-world moving objects, spanning more than 560 object classes and 80+ motion patterns. A key aspect of GOT-10k is its one-shot evaluation protocol, which requires trackers to be trained exclusively on the designated training split, with 170 videos reserved for testing.

\vspace{-9pt}

\paragraph{TrackingNet~\cite{muller2018trackingnet}} is a large-scale tracking dataset that covers a wide selection of object classes in broad and diverse contexts in the wild. It has a total of 30,643 videos split into 30,132 training videos and 511 testing videos. 

\vspace{-9pt}

\paragraph{NFS~\cite{kiani2017nfs}} consists of 100 videos with a total of 380k frames captured with higher frame rate (240 FPS) cameras from real-world scenarios. We use the 30 FPS version of the data with artificial motion blur following other VOT works.

\vspace{-9pt}

\paragraph{OTB100~\cite{wu2015otb}} is one of the earliest visual tracking benchmarks that annotated sequences with attribute tags. It contains 100 sequences with an average length of 590 frames.


\begin{figure}[t]
    \centering
    \begin{subfigure}[b]{0.49\columnwidth}
        \centering
        \includegraphics[width=1.05\textwidth]{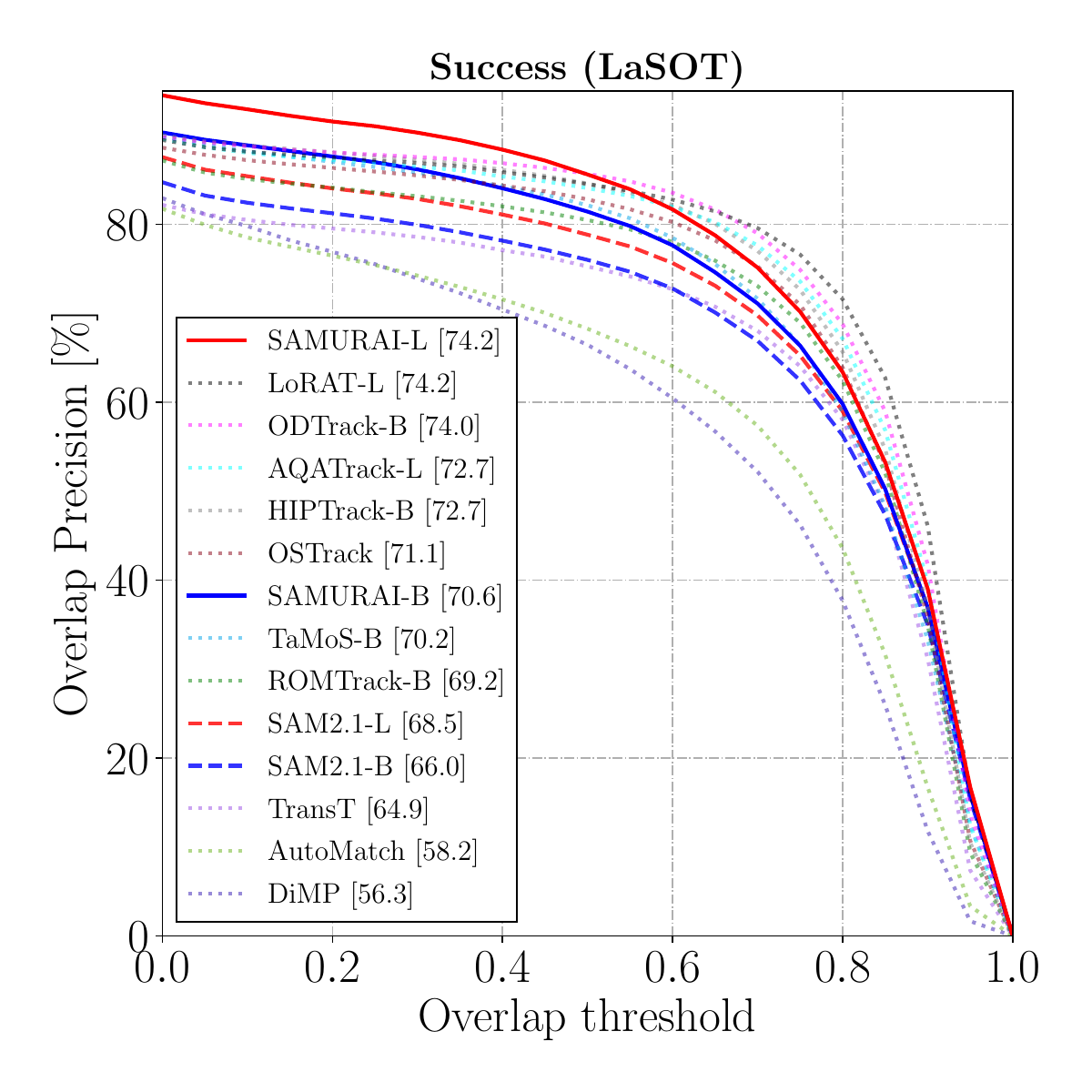}
    \end{subfigure}
    \begin{subfigure}[b]{0.49\columnwidth}
        \centering
        \includegraphics[width=1.05\textwidth]{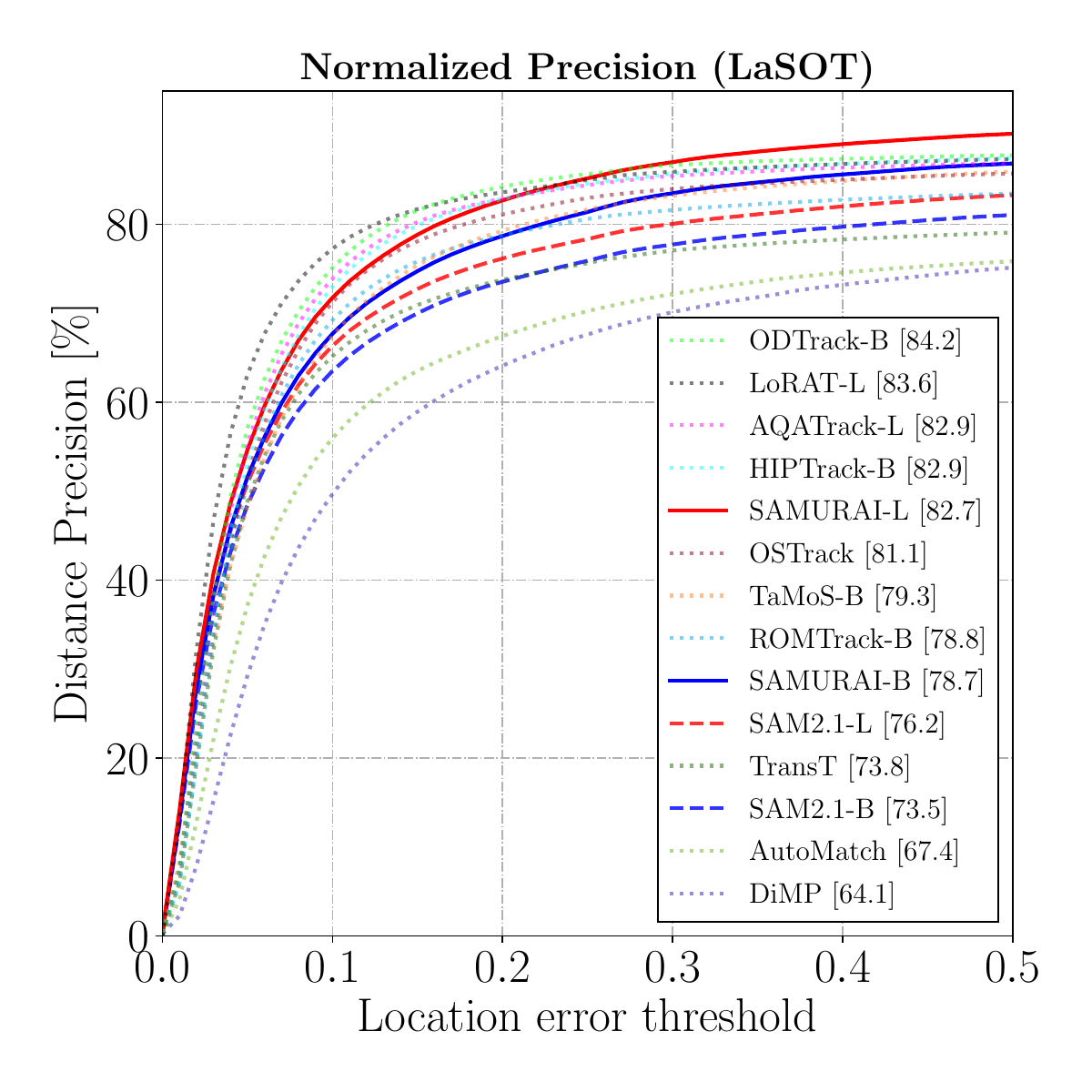}
    \end{subfigure}
    \\
    \begin{subfigure}[b]{0.49\columnwidth}
        \centering
        \includegraphics[width=1.05\textwidth]{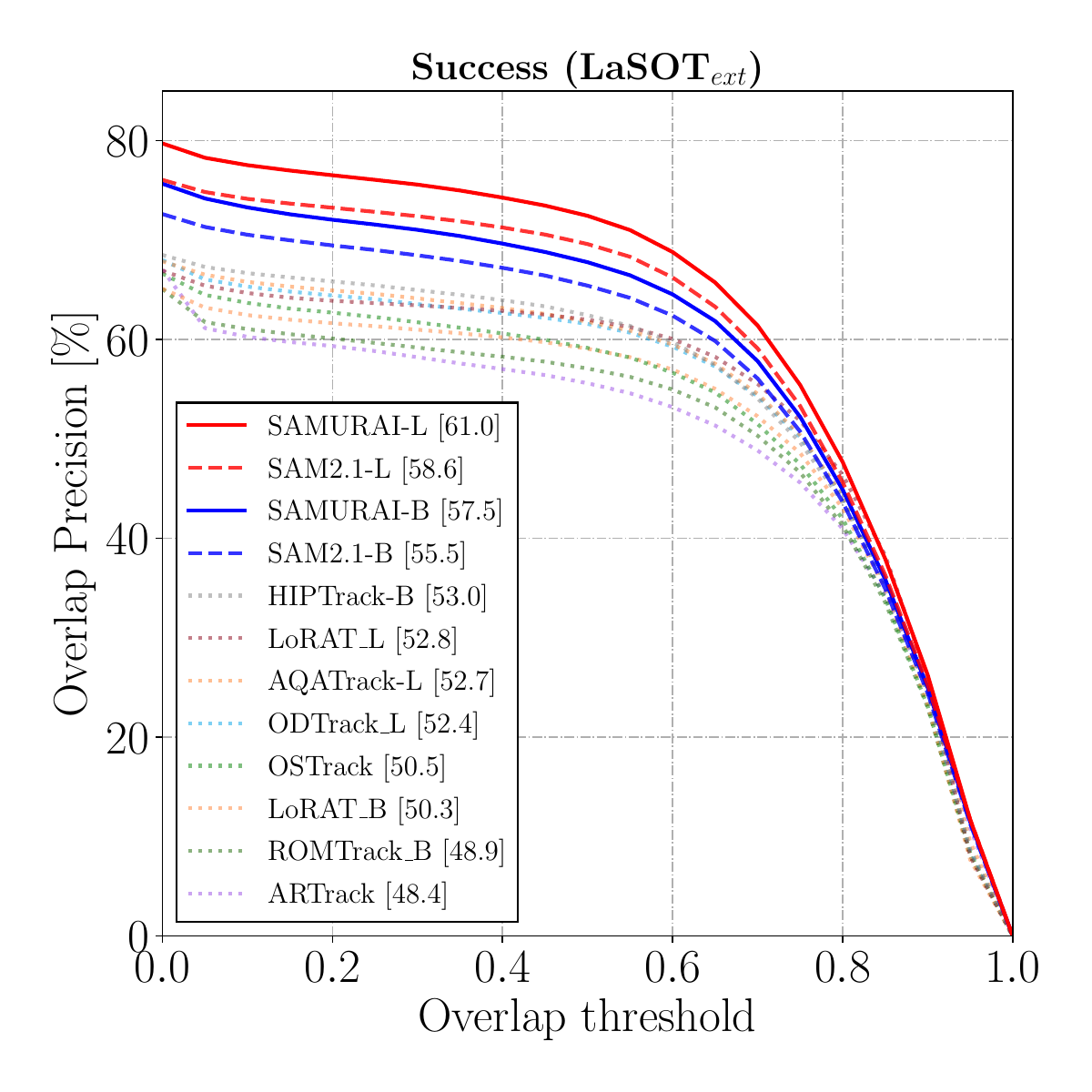}
    \end{subfigure}
    \begin{subfigure}[b]{0.49\columnwidth}
        \centering
        \includegraphics[width=1.05\textwidth]{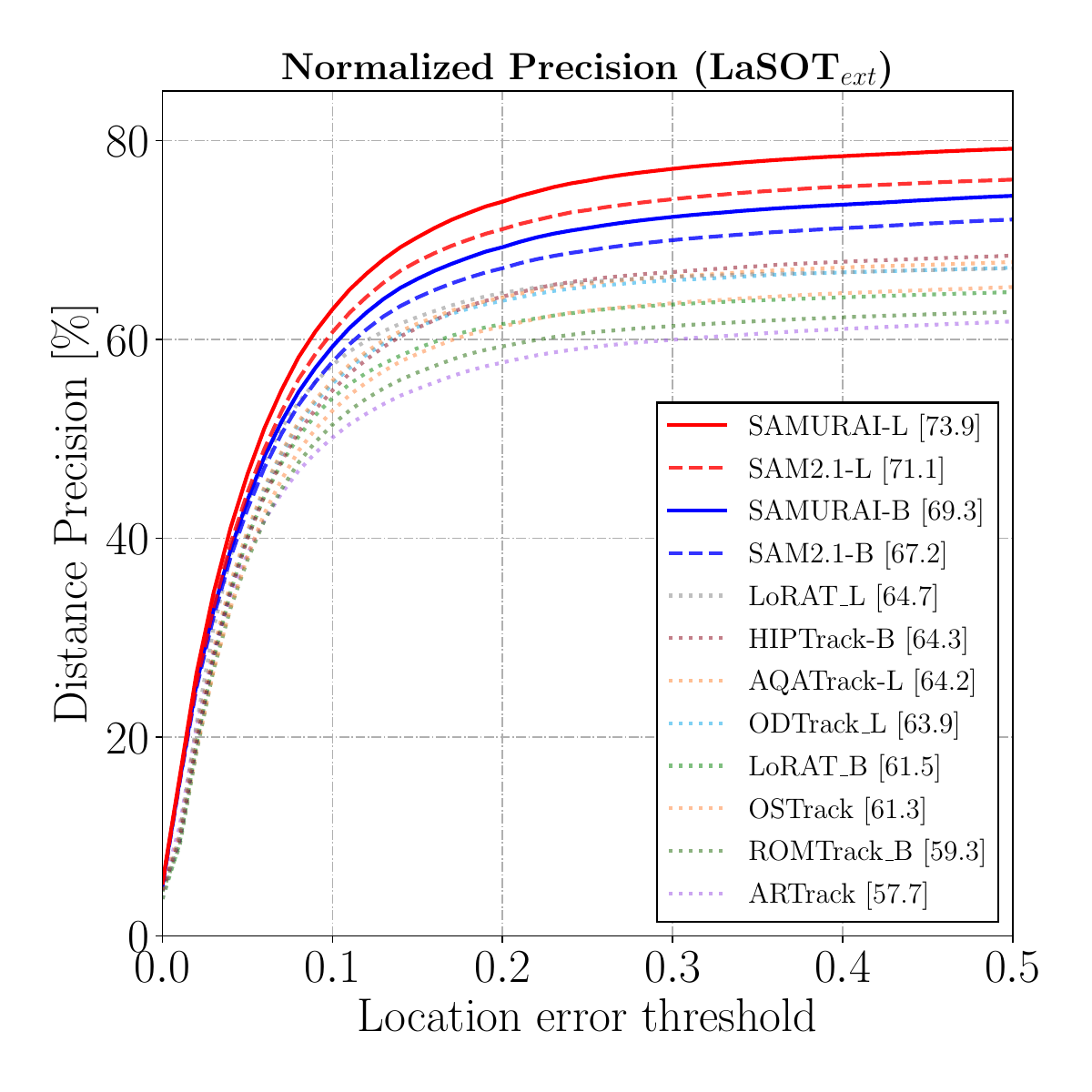}
    \end{subfigure}
       \caption{SUC and P$_{\text{norm}}$ plots of LaSOT and LaSOT$_\text{ext}$.}
       \label{fig:lasot_plots}
\end{figure}

\subsection{Quantitative Results}

\paragraph{Results on \textbf{LaSOT} and \textbf{LaSOT$_{\text{ext}}$}.}

Table \ref{tab:lasot} presents the visual object tracking results on the LaSOT and LaSOT$_{\text{ext}}$ datasets. Our method, ~\model, demonstrates significant improvements over both the zero-shot and supervised methods on all three metrics (shown in Figure \ref{fig:lasot_plots}). Although the supervised VOT method such as \cite{zheng2024odtrack, lin2025lorat} show quite impressive results, the zero-shot~\model~in contrast show its great generalization ability with comparalbe zero-shot performance. Furthermore, all~\model~models surpass the state-of-the-art on all metrics on LaSOT$_{\text{ext}}$.

\begin{table}
    \centering
    \caption{Visual object tracking results on AUC (\%) of our proposed method with state-of-the-art methods on \textbf{TrackingNet}~\cite{muller2018trackingnet}, \textbf{NFS}~\cite{kiani2017nfs}, and \textbf{OTB100}~\cite{wu2015otb} datasets. \textbf{Bold} represents the best while \underline{underline} represents the second.}
    \resizebox{0.95\linewidth}{!}{
    \begin{tabular}{l|ccc}
        \toprule
        Trackers & TrackingNet & NFS & OTB100 \\
        \midrule
        \textit{Supervised method} & & & \\
         DiMP$_{288}$ \cite{danelljan2020dimp} & 74.0 & 61.8 & -\\
         TransT$_{256}$ \cite{chen2021transt} & 81.4 & 65.7 & -\\
         STARK$_{320}$ \cite{yan2021stark} & 82.0 & - & 68.5 \\
         KeepTrack \cite{mayer2021keeptrack} & - & 66.4 & 70.9\\
         AiATrack$_{320}$ \cite{gao2022aiatrack} & 82.7 & 67.9 & 69.6 \\
         OSTrack$_{384}$ \cite{ye2022ostrack} & 83.9 & 66.5 & 55.9 \\
         SeqTrack-B$_{384}$ \cite{chen2023seqtrack} & 83.9 & 66.7 & - \\
         HIPTrack-B$_{384}$  \cite{cai2024hiptrack}& 84.5 & \underline{68.1} & 71.0\\
         AQATrack-L$_{384}$ \cite{xie2024aqatrack} & 84.8 & - & - \\
         LoRAT-L$_{224}$ \cite{lin2025lorat} & \underline{85.0} & 66.0 & \textbf{72.3}\\
         \midrule

         \textit{Zero-shot method} & & & \\
         \rowcolor{aliceblue}SAMURAI-L (Ours) & \textbf{85.3} & \textbf{69.2} & \underline{71.5}\\
         \bottomrule
    \end{tabular}}
    \label{tab:tuno}
\end{table}

\begin{table}[t]
    \caption{Ablation on the effectiveness of the proposed modules.}
    \label{tab:abla_module}
    \small
    \centering
    \resizebox{0.85\linewidth}{!}{
    \begin{tabular}{cc|ccc}
    \toprule
        Motion & Memory & AUC(\%) & P$_{\text{norm}}$(\%) & P(\%) \\ \midrule
        $\times$ & $\times$ & 68.32 & 76.16 & 73.59\\
        $\checkmark$ & $\times$ & 70.81 & 78.87 & 76.47\\
        $\times$ & $\checkmark$ & 72.67 & 80.67 & 78.23\\
        \rowcolor{aliceblue}$\checkmark$ & $\checkmark$ & \textbf{74.23} & \textbf{82.69} & \textbf{80.21} \\
    \bottomrule
    \end{tabular}
    }
\end{table}

\begin{table}[t]
    \caption{Ablation on the sensitivity of the motion weight $\alpha_{kf}$.}
    \label{tab:abla_motion_weight}
    \small
    \centering
    \resizebox{0.75\linewidth}{!}{
    \begin{tabular}{c|ccc}
    \toprule
        $\alpha_{kf}$ & AUC($\%$) & P$_{\text{norm}}$($\%$) & P($\%$) \\ \midrule
        0.00 &  72.67 & 80.67 & 78.23\\
        \rowcolor{aliceblue} 0.15 &  \textbf{74.23} & \textbf{82.69} & \textbf{80.21}\\
        0.25 & 73.76 & 81.86 & 79.53\\
        0.50 & 72.92  & 80.49 & 78.34\\
    \bottomrule
    \end{tabular}
    }
\vspace{-6pt}
\end{table}

\begin{table*}[t]
    \centering
        \caption{Visual object tracking results of the proposed SAMURAI compare to the baseline SAM-based tracking method.}
    \vspace{-6pt}
    \resizebox{0.83\linewidth}{!}{
    \begin{tabular}{l|ccc|ccc}
        \toprule
        \multirow{2}{*}{Trackers} &  \multicolumn{3}{c|}{LaSOT} & \multicolumn{3}{c}{LaSOT$_{\text{ext}}$}  \\ 
         & AUC(\%) & P$_{\text{norm}}(\%)$ & P(\%) & AUC(\%) & P$_{\text{norm}}(\%)$ & P(\%)  \\ \midrule
         SAM2.1-T \cite{ravi2024sam2} & 66.70 &  73.70 & 71.22 & 52.25 & 62.03 & 60.30\\ 
        \rowcolor{aliceblue} SAMURAI-T & \textbf{69.28} \textcolor{OliveGreen}{\textbf{(+2.58)}} & \textbf{76.39} \textcolor{OliveGreen}{\textbf{(+2.69)}} & \textbf{73.78} \textcolor{OliveGreen}{\textbf{(+2.56)}} & \textbf{55.13} \textcolor{OliveGreen}{\textbf{(+2.88)}} & \textbf{65.60} \textcolor{OliveGreen}{\textbf{(+2.57)}} & \textbf{63.72} \textcolor{OliveGreen}{\textbf{(+3.42)}}\\         
        SAM2.1-S \cite{ravi2024sam2} & 66.47 & 73.67 & 71.25 & 56.11 & 67.57 & 65.81\\ 
         \rowcolor{aliceblue}SAMURAI-S &  \textbf{70.04} \textcolor{OliveGreen}{\textbf{(+3.57)}} & \textbf{77.55} \textcolor{OliveGreen}{\textbf{(+3.88)}} & \textbf{75.23} \textcolor{OliveGreen}{\textbf{(+3.98)}} & \textbf{57.99} \textcolor{OliveGreen}{\textbf{(+1.88)}} & \textbf{69.60} \textcolor{OliveGreen}{\textbf{(+2.03)}} & \textbf{67.73} \textcolor{OliveGreen}{\textbf{(+1.92)}} \\ 
         SAM2.1-B \cite{ravi2024sam2} & 65.97 & 73.54 & 70.96 & 55.51 & 67.17 & 64.55 \\ 
         \rowcolor{aliceblue}SAMURAI-B &\textbf{70.65} \textcolor{OliveGreen}{\textbf{(+4.68)}} & \textbf{78.69} \textcolor{OliveGreen}{\textbf{(+4.15)}} & \textbf{76.21} \textcolor{OliveGreen}{\textbf{(+5.25)}} & \textbf{57.48} \textcolor{OliveGreen}{\textbf{(+1.97)}} & \textbf{69.28} \textcolor{OliveGreen}{\textbf{(+2.11)}} & \textbf{67.09} \textcolor{OliveGreen}{\textbf{(+2.54)}}\\
         SAM2.1-L \cite{ravi2024sam2} & 68.54 & 76.16 & 73.59 & 58.55& 71.10& 68.83\\ 
         \rowcolor{aliceblue} SAMURAI-L  & \textbf{74.23} \textcolor{OliveGreen}{\textbf{(+5.69)}} & \textbf{82.69} \textcolor{OliveGreen}{\textbf{(+6.53)}}& \textbf{80.21} \textcolor{OliveGreen}{\textbf{(+6.62)}}& \textbf{61.03} \textcolor{OliveGreen}{\textbf{(+2.48)}}& \textbf{73.86} \textcolor{OliveGreen}{\textbf{(+2.76)}}& \textbf{72.24} \textcolor{OliveGreen}{(\textbf{+3.41})}\\
         \bottomrule
    \end{tabular}
    }
    \label{tab:ablation-baseline}
\end{table*}
\begin{table*}[t]
    \centering
    \caption{Attribute-wise AUC(\%) Results for LaSOT \cite{fan2019lasot} and LaSOT$_{\text{ext}}$ \cite{fan2021lasotext}.}
    \vspace{-6pt}
    \label{tab:my_label}
    \resizebox{0.97\linewidth}{!}{
    \begin{tabular}{l|cccccccccccccc}
    \toprule
    \multirow{2}{*}{Trackers} & \multicolumn{14}{c}{LaSOT}  \\ 
                                 & ARC & BC & CM & DEF & FM & FOC & IV & LR & MB & OV & POC & ROT & SV & VC  \\ \midrule
         SAM2.1-B \cite{ravi2024sam2} & 64.7 & 62.8 & 67.7 & 67.1 & 56.1 & 57.6 & 63.0 & 55.4 & 67.1 & 56.2 & 64.6 & 62.8 & 65.5  & 59.8 \\
         \rowcolor{aliceblue}\model-B & 69.6 & 68.0 & 73.1 & 72.0 & 62.5 & 63.0 & 69.6 & 63.2 & 70.2 & 64.5 & 69.1 & 68.0 & 70.3  & 64.1 \\ 
         \rowcolor{aliceblue}\% Gain & \textcolor{OliveGreen}{+7.6\%} & \textcolor{OliveGreen}{+8.3\%} & \textcolor{OliveGreen}{+8.0\%} & \textcolor{OliveGreen}{+7.3\%} & \textcolor{OliveGreen}{+11.4\%} & \textcolor{OliveGreen}{+9.4\%} & \textcolor{OliveGreen}{+10.5\%} & \textcolor{OliveGreen}{+14.1\%} & \textcolor{OliveGreen}{+4.6\%} & \textcolor{OliveGreen}{+14.8\%} & \textcolor{OliveGreen}{+7.0\%} & \textcolor{OliveGreen}{+8.3\%} & \textcolor{OliveGreen}{+7.3\%} & \textcolor{OliveGreen}{+7.2\%} \\    
         \midrule
         SAM2.1-L \cite{ravi2024sam2} & 67.3 & 64.3 & 69.4 & 70.8 & 58.4 & 59.3 & 63.9 & 59.7 & 67.8 & 61.9 & 68.0 & 67.2 & 68.1 & 61.1 \\
         \rowcolor{aliceblue}\model-L& 73.1 & 69.5 & 77.0 & 75.7 & 63.9 & 66.8 & 72.8 & 67.6 & 73.8 & 70.4 & 72.8 & 72.7 & 73.7 & 71.4 \\ 
         \rowcolor{aliceblue}\% Gain & \textcolor{OliveGreen}{+8.9\%} & \textcolor{OliveGreen}{+8.1\%} & \textcolor{OliveGreen}{+11.0\%} & \textcolor{OliveGreen}{+6.9\%} & \textcolor{OliveGreen}{+9.4\%} & \textcolor{OliveGreen}{+12.7\%} & \textcolor{OliveGreen}{+14.3\%} & \textcolor{OliveGreen}{+13.2\%} & \textcolor{OliveGreen}{+8.0\%} & \textcolor{OliveGreen}{+7.8\%} & \textcolor{OliveGreen}{+7.1\%} & \textcolor{OliveGreen}{+8.2\%} & \textcolor{OliveGreen}{+9.2\%} & \textcolor{OliveGreen}{+16.8\%} \\ \midrule
       \multirow{2}{*}{Trackers} & \multicolumn{14}{c}{LaSOT$_{\text{ext}}$}  \\ 
                                 & ARC & BC & CM & DEF & FM & FOC & IV & LR & MB & OV & POC & ROT & SV & VC  \\ \midrule
    SAM2.1-B & 53.4 & 49.3 & 58.6 & 75.4 & 42.1 & 42.5 & 69.5 & 45.3 & 42.6 & 46.1 & 56.3 & 61.6 & 54.4 & 57.1 \\ 
    \rowcolor{aliceblue}\model-B & 54.8 & 52.5 & 67.8 & 73.3 & 45.9 & 45.9 & 67.3 & 47.4 & 43.7 & 48.1 & 56.7 & 62.9 & 56.2 & 61.8 \\ 
    \rowcolor{aliceblue}\% Gain & \textcolor{OliveGreen}{+2.6\%} & \textcolor{OliveGreen}{+6.5\%} & \textcolor{OliveGreen}{+15.7\%} & \textcolor{OliveGreen}{+10.6\%} & \textcolor{OliveGreen}{+9.0\%} & \textcolor{OliveGreen}{+8.0\%} & \textcolor{BrickRed}{-2.3\%} & \textcolor{OliveGreen}{+4.6\%} & \textcolor{OliveGreen}{+2.6\%} & \textcolor{OliveGreen}{+6.1\%} & \textcolor{OliveGreen}{+11.5\%} & \textcolor{OliveGreen}{+2.1\%} & \textcolor{OliveGreen}{+3.3\%} & \textcolor{OliveGreen}{+8.2\%} \\ 
    \midrule
    SAM2.1-L & 56.6 & 53.2 & 62.8 & 75.6 & 46.1 & 47.6 & 71.4 & 48.8 & 47.1 & 50.9 & 60.0 & 63.2 & 57.7 & 61.9 \\ 
   \rowcolor{aliceblue}\model-L~ & 58.4 & 55.4 & 73.1 & 77.5 & 50.7 & 50.9 & 69.0 & 52.1 & 49.4 & 53.3 & 60.9 & 64.7 & 59.9 & 66.6 \\ 
    \rowcolor{aliceblue}\% Gain & \textcolor{OliveGreen}{+3.2\%} & \textcolor{OliveGreen}{+4.0\%} & \textcolor{OliveGreen}{+16.5\%} & \textcolor{OliveGreen}{+5.2\%} & \textcolor{OliveGreen}{+9.9\%} & \textcolor{OliveGreen}{+6.9\%} & \textcolor{BrickRed}{-3.5\%} & \textcolor{OliveGreen}{+6.3\%} & \textcolor{OliveGreen}{+5.4\%} & \textcolor{OliveGreen}{+4.3\%} & \textcolor{OliveGreen}{+3.2\%} & \textcolor{OliveGreen}{+2.6\%} & \textcolor{OliveGreen}{+4.0\%} & \textcolor{OliveGreen}{+7.4\%} \\ 
    \bottomrule
    \end{tabular}
    }
\end{table*}


\vspace{-6pt}

\paragraph{Results on GOT-10k.} Table~\ref{tab:lasot} also presents the visual object tracking results on the GOT-10k dataset. Note that the GOT-10k protocol only allows trackers to be trained using its corresponding train split, as some papers may refer to them as a one-shot method. ~\model-B shows a 2.1\% improvement on AO and 2.9\% on OP$_{0.5}$ over SAM2.1-B while~\model-L shows a 0.6\% improvement on AO and 0.7\% on OP$_{0.5}$. All~\model~models surpass the state-of-the-art on all metrics on GOT-10k.

\vspace{-6pt}

\paragraph{Results on \textbf{TrackingNet}, \textbf{NFS}, and \textbf{OTB100}.} Table~\ref{tab:tuno} presents the visual object tracking results on four widely compared benchmarks. Our zero-shot~\model-L model is comparable to or can surpass the state-of-the-art supervised method on AUC, showcasing the capability of our model on various datasets and generalization ability. 

\subsection{Ablation Studies}

\paragraph{Effect of the Individual Modules.} We demonstrate the effect of the with or without memory selection on various settings in Table~\ref{tab:abla_module}. Both of the proposed modules had a positive impact on the SAM 2 model, while combining both can achieve the best AUC on the LaSOT dataset with an AUC of 74.23\% and P$_\text{norm}$ of 82.60\%. 

\vspace{-6pt}
\paragraph{Effect of the Motion Weights.} We showcase the effect of the weighting of the score of deciding which mask to trust in Table~\ref{tab:abla_motion_weight}. The trade-off between motion score and mask affinity score demonstrates a significant impact on tracking performance. Our experiments reveal that setting the motion weight $\alpha_{motion}=0.2$ yields the best AUC and P$_\text{norm}$ score on the LaSOT dataset, indicating an optimal balance enhances both accuracy and robustness in mask selection.

\begin{figure*}[tbh]
    \centering
     \begin{subfigure}{0.97\textwidth}
        \centering
        \includegraphics[width=0.98\linewidth]{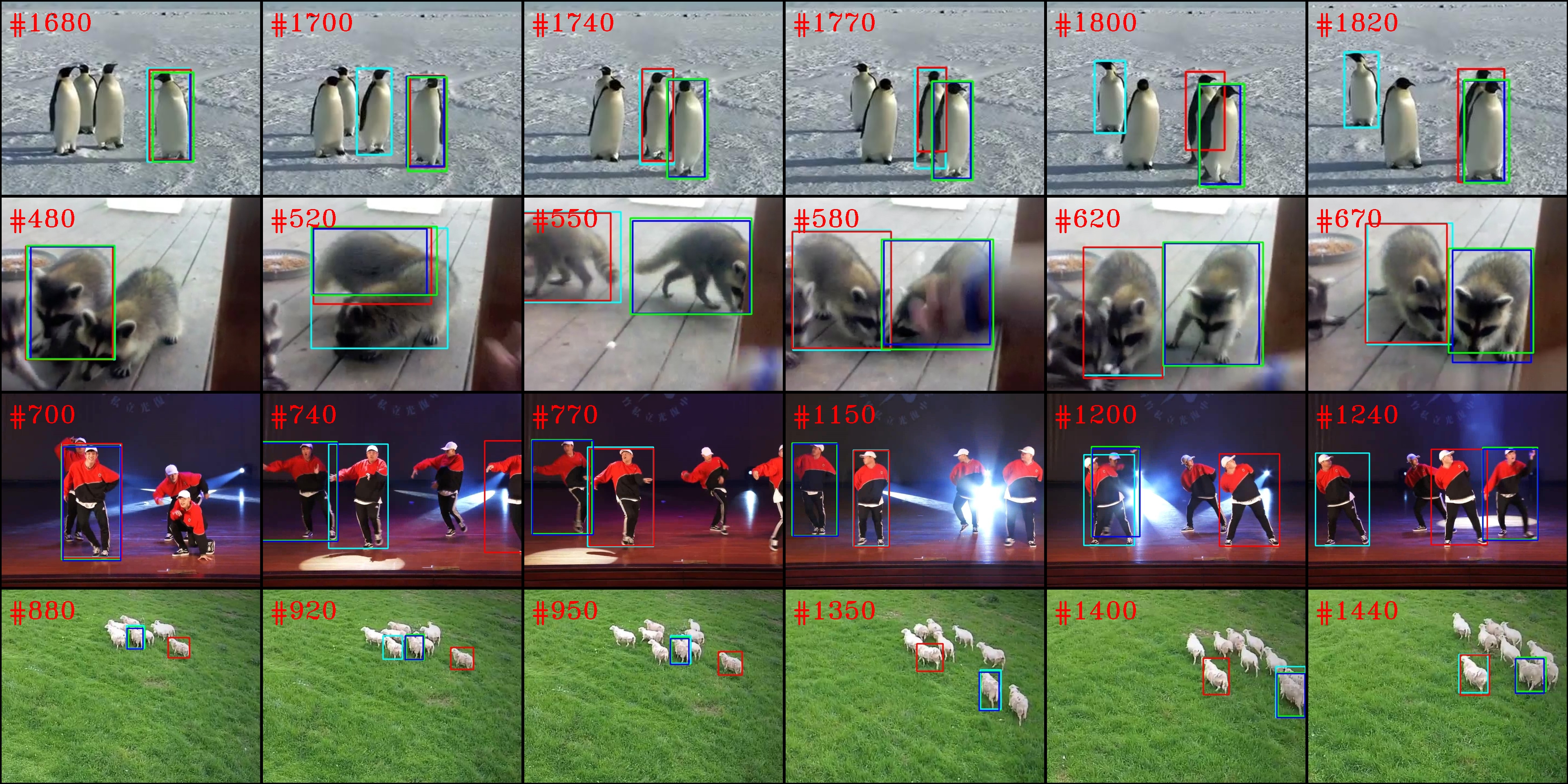}
        \vspace{6pt}
        {\raisebox{3pt}{\tikz{\draw[red,solid,line width=1.5pt](0mm,0mm) -- (5mm,0mm);}} HIPTrack \quad \raisebox{3pt}{\tikz{\draw[cyan,solid,line width=1.5pt](0,0mm) -- (5mm,0mm);}} LoRAT \quad \raisebox{3pt}{\tikz{\draw[blue,solid,line width=1.5pt](0,0mm) -- (5mm,0mm);}}~\model~(Ours) \quad \raisebox{3pt}{\tikz{\draw[green,solid,line width=1.5pt](0,0mm) -- (5mm,0mm);}} GT}  
    \end{subfigure}
    \begin{subfigure}{0.97\textwidth}
        \centering
        \includegraphics[width=0.98\linewidth]{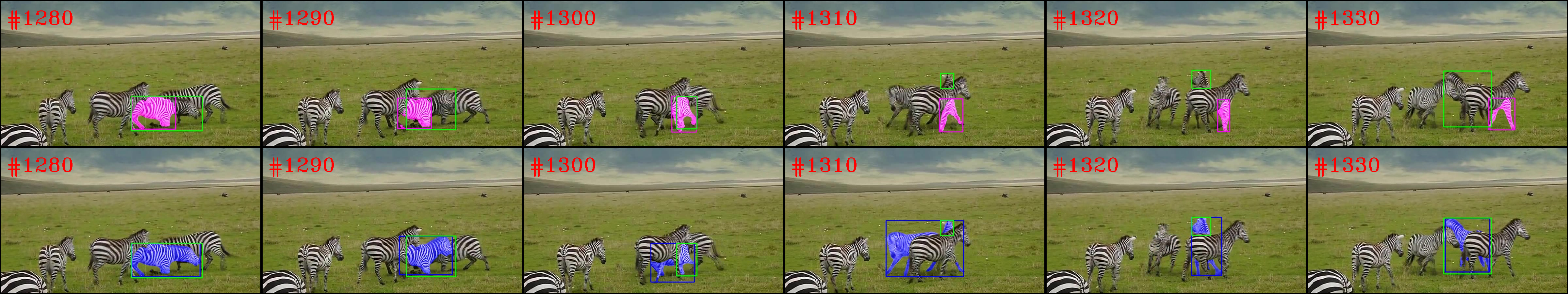}
        {\raisebox{3pt}{\tikz{\draw[magenta,solid,line width=1.5pt](0,0mm) -- (5mm,0mm);}} SAM2.1~(Baseline) \quad \raisebox{3pt}{\tikz{\draw[blue,solid,line width=1.5pt](0,0mm) -- (5mm,0mm);}}~\model~(Ours) \quad \raisebox{3pt}{\tikz{\draw[green,solid,line width=1.5pt](0,0mm) -- (5mm,0mm);}} GT}  
    \end{subfigure}
    \caption{Visualization of tracking results comparing \model with existing methods. (Top) Conventional VOT methods often struggle in crowded scenarios where the target object is surrounded by objects with similar appearances. (Bottom) The baseline SAM-based method suffers from fixed-window memory composition, leading to error propagation and reduced overall tracking accuracy due to ID switches.}
    \label{fig:viz}
\end{figure*}
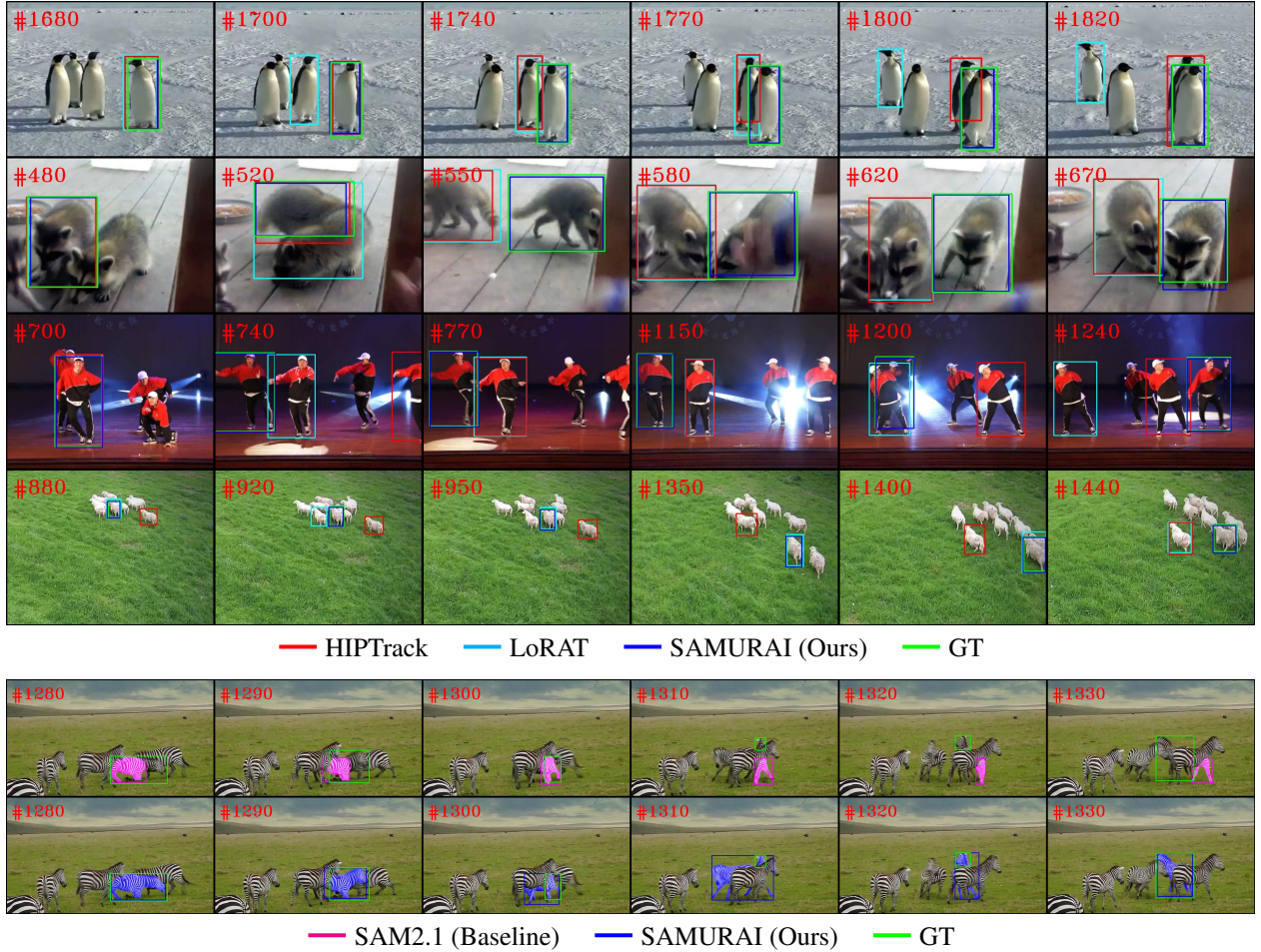

\vspace{-1em}

\paragraph{Baseline Comparison.} To demonstrate the effectiveness of the proposed motion modeling and motion-aware memory selection mechanism in SAMURAI, we conduct a detailed apple-to-apple comparison of the SAM 2 \cite{ravi2024sam2} at all of the backbone variations on LaSOT and LaSOT$_{\text{ext}}$. The baseline SAM 2 employs the original memory selection and directly predicts the mask with the highest IoU score. Table~\ref{tab:ablation-baseline} shows that the proposed method consistently improves upon the baseline with a significant margin on all three metrics, which underscores the robustness and generalization of our approach across different model configurations.

\vspace{-1em}

\paragraph{Attribute-Wise Performance Analysis.}  We analysis the LaSOT and LaSOT$_{\text{ext}}$ based on the 14 attributes defined in \cite{fan2019lasot, fan2021lasotext}. In Table~\ref{tab:my_label}, \model~shows consistent success in improving upon the original baseline across all attributes in both datasets but the label IV (Illumination Variation) label on LaSOT$_{\text{ext}}$. By considering motion scoring, the performance gains on attributes like CM (Camera Motion) and FM (Fast Motion) are the largest among the rest, the~\model~has a $16.5\%$ and $9.9\%$ gain on CM and FM respectively from LaSOT$_{\text{ext}}$ dataset which is considered one of the most challenging datasets in VOT. Furthermore, the occlusion-related attributes like FOC (Full Occlusion) and POC (Partial Occlusion) also greatly benefited from the proposed motion-aware instance-level memory selection, which showed steady improvement across all model variants and datasets. These findings suggest that the~\model~incorporates simple motion estimation to better account for global camera or rapid object movements for better tracking.

\paragraph{Runtime Analysis.} The incorporation of the motion modeling and an enhanced memory selection method into our tracking framework introduces minimal computational overhead, and the runtime measurements conducted on one NVIDIA RTX 4090 GPU remain consistent with the baseline model. 

\subsection{Qualitative Results}

Qualitative comparison between~\model~and other methods \cite{ravi2024sam2, cai2024hiptrack, lin2025lorat} are shown in Figure \ref{fig:viz}. \model~demonstrates superior visual object tracking results in scenes where multiple objects with similar appearances are present in the video. The short-term occlusions in these examples make it challenging for existing VOT methods to predict or localize the same object consistently over time. Furthermore, the comparison between \model~and the original baseline with visualized masks showcases the improvement gained by adding the motion modeling and memory selection modules, the predicted masks are not always a reliable source to serve as memory therefore having a systematic way of deciding which to trust is valuable. These enhancements benefit the existing framework by providing better guidance for visual tracking without the need to retrain the model or fine-tune it.

\section{Conclusion}

We present~\model, a visual object tracking framework built on top of the segment-anything model by introducing the motion-based score for better mask prediction and memory selection to deal with self-occlusion and abrupt motion in crowded scenes. The proposed modules show consistent improvement on all variations of the SAM models across multiple VOT benchmarks on all metrics. This method does not require re-training nor fine-tuning while demonstrating robust performance on multiple VOT benchmarks with the capability of real-time online inferences.




\newpage
{
    \small
    \bibliographystyle{ieeenat_fullname}
    \bibliography{main}
}

\end{document}